\documentclass[10pt, journal, compsoc]{IEEEtran}
\newif\ifpeerreview

\peerreviewfalse
\usepackage{packages}
\usepackage{subcaption}

\crefname{section}{Sec.}{Secs.}
\Crefname{section}{Sec.}{Secs.}
\crefname{figure}{Fig.}{Figs.}
\Crefname{figure}{Fig.}{Figs.}
\crefname{equation}{Eq.}{Eqs.}
\Crefname{equation}{Eq.}{Eqs.}
\crefname{table}{Tab.}{Tabs.}
\Crefname{table}{Tab.}{Tabs.}

\usepackage{caption}
\DeclareCaptionFont{ieeebold}{\sffamily\bfseries}
\DeclareCaptionFont{ieeesans}{\sffamily}

\newcommand{\us}{\text{us}}
\newcommand{\ssub}{\text{us}}

\DeclareSIUnit{\fps}{fps}
\DeclareSIUnit{\rpm}{rpm}
\DeclareSIUnit{\line}{lines}
\DeclareSIUnit{\mil}{M}

\renewcommand{\paragraph}[1]{\vspace{0.2em}\noindent{\bf #1}}

\newcommand{\igkiou}[1]{{}}

\definecolor{darkgreen}{RGB}{0, 153, 0} % Define the color #009900
\definecolor{Maroon}{rgb}{0.5, 0.0, 0.0} % Example RGB definition

\definecolor{Maroon}{rgb}{0.5, 0.0, 0.0} % Example RGB definition

\title{Structured light with a million light planes per second}

\author{ 
Dhawal Sirikonda$^{\dagger}$, Praneeth Chakravarthula$^{\ddagger}$, Ioannis Gkioulekas$^{\S}$, Adithya K Pediredla$^{\dagger}$ \\ $^{\dagger}$Dartmouth College, $^{\ddagger}$UNC Chapel Hill, $^{\S}$Carnegie Mellon University}

\begin{document}
    \IEEEtitleabstractindextext{%
    
    \begin{abstract}
We introduce a structured light system that enables full-frame 3D scanning at speeds of \SI{1000}{\fps}, four times faster than the previous fastest systems. Our key innovation is the use of a custom acousto-optic light scanning device capable of projecting two million light planes per second. Coupling this device with an event camera allows our system to overcome the key bottleneck preventing previous structured light systems based on event cameras from achieving higher scanning speeds---the limited rate of illumination steering. Unlike these previous systems, ours uses the event camera's full-frame bandwidth, shifting the speed bottleneck from the illumination side to the imaging side. To mitigate this new bottleneck and further increase scanning speed, we introduce adaptive scanning strategies that leverage the event camera's asynchronous operation by selectively illuminating regions of interest, thereby achieving effective scanning speeds an order of magnitude beyond the camera's theoretical limit. 
\end{abstract}
     }
    
    % Make Title
    \ifpeerreview \linenumbers \linenumbersep 15pt\relax
    \author{\paperID\IEEEcompsocitemizethanks{\IEEEcompsocthanksitem This paper is under review for ICCP 2025 and the PAMI special issue on computational photography. Do not distribute.}}
    \markboth{Anonymous ICCP 2025 submission ID \paperID}%
    {} \fi
    
    \twocolumn[{
    \renewcommand\twocolumn[1][]{#1}
    \maketitle
    }]
    
    \IEEEraisesectionheading{ \section{Introduction}\label{sec:introduction} }

\IEEEPARstart{3}{D} scanning systems acquire the detailed geometry
of real-world objects, using methods such as time of flight, structured light, or photogrammetry. Though such systems are already commonplace for static scenes, nowadays there is an increasing demand for 3D scanning systems that can operate on dynamic scenes and provide real-time 3D information for accurate analysis of moving objects. For instance, such systems can enhance navigation and safety in autonomous driving and robotics, enable interactive experiences in virtual and augmented reality, or allow for inspection of fast-moving objects in industrial manufacturing. These applications require \emph{fast} 3D scanning that minimizes motion blur and other motion-induced artifacts.

Fast 3D scanning necessitates hardware for both high-speed illumination (e.g., using Galvo mirror systems or MEMS mirror projectors), and high-speed imaging (e.g., using single-photon avalanche diodes or event sensors). It also requires efficient algorithms to process and filter data rapidly, handling tasks such as noise reduction, data fusion, and reconstruction. In scanning systems that utilize multiple sensors, synchronization also plays a crucial role in ensuring coherent data capture and integration from different sources. 
Unfortunately, fast 3D scanning methods are limited to rates below \qty{1000}{\fps}, primarily due to constraints imposed by the scanning speeds of existing light scanning systems: Though fast imaging sensors such as event cameras exist, light scanning systems are limited to scan rates of a few \SI{}{\kilo \hertz}, falling well short of fully exploiting the fast sensor readout.

In this work, we push the envelope of achievable 3D scanning speeds, by
designing an ultra-fast structured light (SL) system that combines an acousto-optic (AO) light scanning device and an event camera. The AO device comprises a pulsed laser and an ultrasonic transducer. The transducer generates an ultrasonic wave that sculpts traveling virtual cylindrical lenses in a water medium. These lenses focus the laser light onto a line. By steering these cylindrical lenses at \SI{}{\mega \hertz} rates, we can scan the line over a 3D scene imaged by the event camera, enabling structured light scanning. The AO device uses a design that results in an order-of-magnitude cost reduction relative to prior such devices \citep{pediredla2023megahertz}.

\begin{figure*}[t]
    \begin{minipage}{\linewidth}
        \centering
        \includegraphics[width=\linewidth]{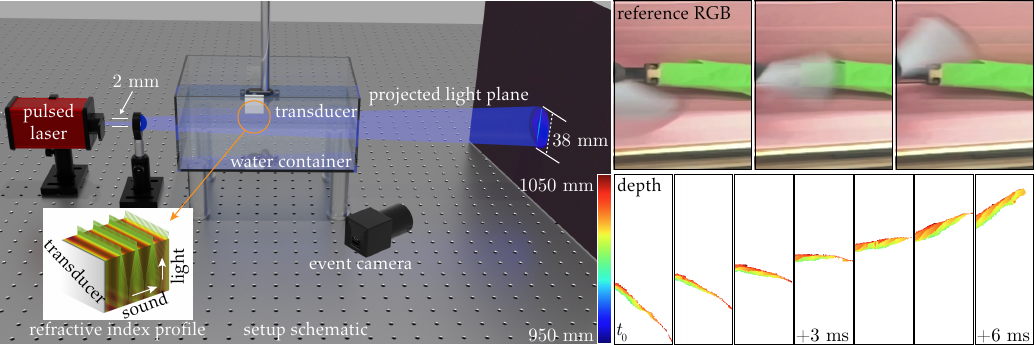}
    \end{minipage}
    \caption{We present a structured light technology that combines acousto-optic light steering with an event camera for high-speed full-frame scanning. Left: Schematic of our setup. We use an ultrasonic transducer to sculpt virtual gradient-index (GRIN) cylindrical lenses in water {(the inset shows the refractive index profile focusing light)}. Coupling this setup with pulsed illumination, we can scan the imaged scene with a light plane at speeds three orders of magnitude faster than those of previous light scanning methods, allowing structured light operation at the camera's full-frame bandwidth. Top-right: RGB images, captured at \qty{240}{\fps}, of a scene comprising a fan that rotates at \qty{1800}{\rpm}. The images correspond to the approximate positions of the depth scans below. Bottom-right: Reconstructed depth frames of the rotating fan, scanned at \SI{1000}{\fps}.}
    \label{fig:schematic}
    \label{fig:teaser}
\end{figure*}

Our AO device achieves light scanning rates that are more than three orders of magnitude higher than the frame rate of the event camera. 
Thus, when used for structured light, our device shifts the bottleneck for achieving higher frame rates from the speed of light scanning system to the readout bandwidth of imaging sensors. To further improve speed beyond this bottleneck, we demonstrate an adaptive scanning method inspired by \citet{muglikar2021event} that illuminates only regions of interest. Doing so allows capture $10\times$ faster than the theoretical full-frame limit of the event sensor. We demonstrate these capabilities experimentally by building a benchtop prototype, and using it to scan a variety of static and dynamic scenes.\footnote{Porject Page: \url{https://dartmouth-risc-lab.github.io/aosl/}}

    \section{Prior work}\label{sec:prior_work}

\paragraph{Light scanning.} Light scanning is a core component of active imaging technologies, including lidar \cite{yang2023development}, structured light \cite{gupta2011structured,sundar2022single, ichimaru2025neural, qiao2024depth}, light-transport probing \cite{o2012primal,o2015homogeneous,achar2017epipolar}, motion contrast 3D \cite{matsuda2015mc3d}, light curtains \cite{chan2022holocurtains,wang2018programmable}, slope-disparity gating \cite{ueda2019slope,kubo2019programmable,chandran2022slope}, and non-line-of-sight imaging \cite{lindell2019wave,liu2019non,o2018confocal,pediredla2019snlos,xin2019theory}. We can broadly classify the light scanning methods in these technologies into mechanical and non-mechanical. Mechanical methods require moving parts such as rotating prisms \cite{abdo2019spatial,wang2019super} or mirrors \cite{kim2004adaptive,mccarthy2013kilometer}, and micro-electromechanical systems (MEMS) \cite{rodriguez2018multi,smith2017single}. These methods are slow due to mechanical inertia. Non-mechanical methods require no moving parts, and include acousto-optic (AO) \cite{gavryusev2019dual,xu2017tunable} and electro-optic (EO) \cite{sun2001polymeric} devices, liquid crystal devices (LCDs) \cite{shang2015electrically,wang2016agile}, and optical phased arrays (OPAs) \cite{hsu2020review,poulton20208192}. LCDs are the slowest of these methods due to long settling times. OPAs are the fastest and can scan light at even \SI{}{\giga \hertz} rates, but have low angular resolutions, and thus are unsuitable for structured light. AO and EO devices provide a good middle ground, as we discuss next.

\begin{table}[t]
    \centering
    \caption{Scan speed comparison against recent works.}\label{tab:prior}
    \begin{tabularx}{\linewidth}{@{} >{\hsize=1.3\hsize}X >{\hsize=0.85\hsize\centering\arraybackslash}X >{\hsize=0.85\hsize\centering\arraybackslash}X @{}}
    \toprule
    \textbf{Work}               & \textbf{Static}       & \textbf{Dynamic}  \\
    \midrule
    \citet{muglikar2021esl}     & \SI{60}{\fps}       & \SI{60}{\fps}    \\
    \citet{matsuda2015mc3d}     & \SI{60}{\fps}       & \SI{60}{\fps}   \\
    \citet{dashpute2023event}   & \SI{250}{\fps}      & \SI{30}{\fps}   \\
    Ours                        & \textcolor{Maroon}{\SI{1}{\kilo\fps}}   & \textcolor{Maroon}{\SI{1}{\kilo\fps}} \\ 
    \bottomrule
    \end{tabularx}
\end{table}

\paragraph{Light scanning with AO and EO devices.} 
Commercially available AO and EO devices include tunable filters \cite{gottlieb2021acousto}, modulators \cite{beller2022acousto}, frequency shifters \cite{yu2021gigahertz}, and deflectors \cite{jeong2022angular}. AO deflectors can serve as light scanning devices, but are limited to \SI{}{\kilo\hertz} scanning rates due to their reliance on Bragg's diffraction \cite{tsai2013guided}. EO deflectors \cite{liu2019electromechanical,song1996electro} operate on a similar principle, and thus are similarly slow. Alternative AO light scanning devices are tunable acoustic gradient-index (TAG) lenses \cite{kang2020variable,cherkashin2020transversally} and ultrasonically-sculpted virtual optical waveguides \cite{chamanzar2019ultrasonic,karimi2019situ,pediredla2023megahertz}---our technology is an instance of the latter. TAG lenses can change the focus depth of an incident beam at \SI{}{\kilo \hertz} rates, but cannot scan it transversely. 

Most closely related to our work, \citet{pediredla2023megahertz} demonstrated transverse scanning at \SI{}{\mega\hertz} rates using ultrasonically-sculpted virtual optical waveguides, motivating our work. Compared to their scanning system, which phase-modulates the ultrasound for scanning, our system modulates the laser pulse. This adjustment allows us to use a \emph{narrowband} amplifier, which is widely available and cost-effective (USD \num{250} instead of USD \num{23000} for the broadband amplifier in \citet{pediredla2023megahertz}), reducing the system cost.

{Lastly, \citet{hullin2011dynamic} used perturbation of surface waves to dynamically display BRDFs. 
However, their technique is limited to \SI{}{\kilo\hertz} range as high frequency waves experience significant damping.
}
    
\paragraph{Event cameras and 3D sensing.} Event cameras perform sparse readouts of relative intensity changes exceeding a preset threshold, and thus can achieve much larger frame rates than conventional cameras that always perform full-frame readouts. This high speed has made event cameras increasingly popular in computer vision \citep{han2020neuromorphic,tulyakov2021time,tulyakov2022time, qu2025event}. Though extensive research exists for 2D imaging applications of event cameras, such as full-frame reconstruction \citep{kim2008simultaneous,rebecq2016evo,scheerlinck2018continuous,pan2020high,shedligeri2019photorealistic,liu2017high} and deblurring \citep{haoyu2020learning, sun2022event, sun2023event,xu2021motion,zhang2020hybrid,pan2019bringing}, 3D imaging applications are still nascent. \citet{brandli2014adaptive} combined an event camera with a laser line scanner for structured-light 3D scanning, leveraging the fact that swept-plane structured light requires very sparse sensor readouts---only a column per projected light plane. More recent works perform structured light by combining event cameras with laser point projectors \citep{matsuda2015mc3d,martel2018active,muglikar2021esl,muglikar2021event} or digital light processing projectors \citep{leroux2018event,mangalore2020neuromorphic}. In all these works, light scanning is slower than the frame rate of the event camera, and thus the bottleneck preventing faster 3D scanning. We remove this bottleneck by using a laser-line scanning system that combines acousto-optic lens sculpting with a pulsed laser, to achieve megahertz-rate light scanning---three orders of magnitude faster than the theoretical frame-rate limit of an event camera, thus turning the camera into the bottleneck towards faster structured light.     
    
    \begin{figure}
    \centering
    \includegraphics[width=\linewidth]{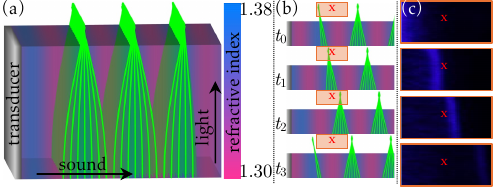}
    \caption{{(a) We show the refractive index profile created by the ultrasonic transducer and how light traveling from bottom to top focuses onto lines. (b) As the refractive index profile moves horizontally over time, the focused lines also move horizontally. (c) Captured images demonstrate the temporal movement of the focused line}}
    \label{fig:ultrafast_line_scanning}
\end{figure}
    \begin{figure}[t]
  \centering
  \includegraphics[width=\linewidth]{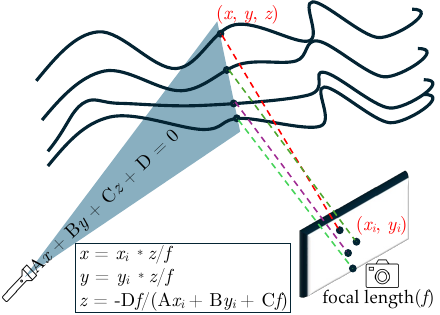}
  \caption{\textbf{Structured light scanning with swept light planes.} The light scanning device sweeps a light plane across the imaged scene. The reflected light triggers events at pixels along a vertical curve on the event camera. For each such pixel, depth is triangulated by intersecting a backprojected ray with the corresponding light plane.}
  \label{fig:structured_light}
\end{figure}
    \begin{figure*}[t]
    \centering
    \includegraphics[width=\linewidth]{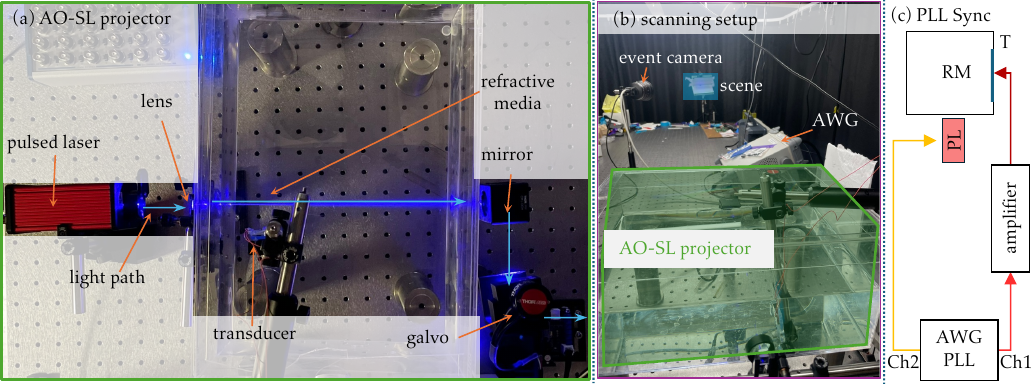}
    \caption{
    \textbf{Hardware implementation of the setup in \cref{fig:schematic}}. (a) AO light scanning device. A pulsed laser (PL) emits a beam that is expanded by a convex lens and directed into a refractive medium (RM). An ultrasonic transducer (T), perpendicular to the beam, generates acoustic waves that sculpt cylindrical lenses in the medium. These lenses focus the light into sweeping light planes, which are then redirected by a fixed mirror onto a Galvo mirror. During acousto-optic scanning, the Galvo mirror remains stationary. Conversely, when using the Galvo mirror for scanning, the transducer and pulse frequency are constant. (b) Scanning setup. The AO device from (a) is shown in green. The scene (in cyan) contains orthogonal planes for calibration. The laser and ultrasound are driven by an arbitrary waveform generator (AWG). (c) Phase-lock-loop (PLL) synchronization of the transducer with the pulsed laser, driven via Channels $1$ and $2$ (Ch$1$, Ch$2$) of the AWG. Channel $1$’s output is amplified using an amplifier (Amp). This PLL synchronization ensures accurate pulse placement with respect to the ultrasound.}
    \label{fig:hardware_prototype}
\end{figure*}
    \section{Method}\label{sec:method}

Structured light (SL) methods triangulate depth by establishing stereo correspondences between an active light projector and a camera. In its simplest form, SL uses the projector to raster scan individual points~\cite{matsuda2015mc3d,muglikar2021esl}. Using the epipolar geometry allows accelerating structured light while maintaining robust correspondences, by projecting a line (light plane in 3D) orthogonal to the epipolar lines. Methods using optical codes such as binary~\cite{posdamer1982surface}, Gray~\cite{caspi1998range}, sinusoidal~\cite{huang2003high}, or XOR~\cite{gupta2011structured} reduce the number of projected light patterns, but are detrimental for the event camera case---approximately half the sensor pixels will be lit for each projected pattern, increasing the number of generated events and thus slowing the scanning rate. 
    
Motivated by these considerations, we opt for a system that uses line projection and an event camera for structured light. We build an acousto-optic line scanning prototype that can project two million lines per second (lps), exceeding the event camera theoretical detection rate ($\approx 10^6$ lps), and being orders of magnitude faster than alternative projection systems used for structured light, including Galvo mirrors~\cite{dashpute2023event,matsuda2015mc3d}, MEMS mirrors~\cite{leroux2018event,mangalore2020neuromorphic}, and laser projectors~\cite{martel2018active}. Below we describe the physical principles governing the operation of the line scanning device, and the algorithmic principles for programmable light scanning. 
       
\paragraph{Ultrasonic sculpting of steerable lenses.} The refractive index of an optical medium is a function of its density.  By controlling a medium's spatial density, we can control its spatially varying refractive index and convert the medium into a gradient-index (GRIN) lens. To this end, we use a planar acoustic transducer to create a pressure wave that sculpts inside a transparent medium (water) a cylindrical GRIN lens traveling at the speed of sound in that medium. Using this virtual cylindrical lens, we can focus a collimated beam of light onto a line that also moves at the speed of sound. We can then use this light beam for structured light.  
If we apply sinusoidal voltage $V(t) = V_{\ssub} \cos\left(2\pi f_{\us}t\right)$ to the planar transducer at $x=0$, the transducer will create ultrasound inducing a traveling pressure wave: 
\begin{equation}
    P(x, y, z, t) = P_{0}  + P_{\ssub}\cos\left(\nicefrac{2\pi}{\lambda_{\us}}x - 2\pi f_{\us}t\right),
    \label{eq:pressure_us}
\end{equation}
where $x, y, z$ are Cartesian coordinates, $t$ is time, $P_{0}$ is the pressure in the medium without ultrasound, $P_{\ssub}$ is proportional to the input voltage amplitude $V_{\ssub}$, and $\lambda_{\us}$ and $f_{\us}$ are the wavelength and frequency (resp.) of the ultrasound output by the transducer. The pressure wave creates a proportional change in the refractive index of the medium, resulting in a time-varying refractive index profile:
\begin{equation}
    \label{eq:ri_equation}
    n(x,y,z, t) =  n_{0} + n_{\ssub} \cos\left(\nicefrac{2\pi}{\lambda_{\us}}x - 2\pi f_{\us}t\right), 
\end{equation}
where $n_{0}$ is the refractive index without ultrasound, and $n_{\ssub}$ is proportional to the pressure amplitude $P_{\ssub}$. We drop $y, z$ as the refractive index does not change along those axes. As \citet{pediredla2023megahertz} explain, each of the the convex lobes of $n(x,t)$---corresponding to $x \in ( k\lambda_{\us} - \nicefrac{\lambda}{2}, k\lambda_{\us} + \nicefrac{\lambda}{2} ),\, k \in \mathbb{Z}$---acts as a cylindrical GRIN lens that travels along the $x$ axis at the speed of ultrasound, % these convex lobes move at the speed of 
$c_\us = f_\us \lambda_\us$.

A light beam passing through the medium will focus to a line---or series of lines if the beam width is larger than the ultrasound wavelength---traveling at speed $c_\us$. Assuming for simplicity that the cylindrical lenses are aberration-free, the resulting intensity at the lens focal plane is
\begin{align}
    I(x, t) = \sum_k \delta(x + k\lambda_\us - c_\us t), 
    \label{eq:continous_intensity_focalplane}
\end{align}
%
% ignoring the constant scale proportional to the illumination power. 
up to a scale factor proportional to the illumination.
\Cref{fig:ultrafast_line_scanning} visualizes the cylindrical lenses, focusing behavior of light rays, and their temporal dynamics.

\paragraph{Programmable control of light planes.}
The traveling cylindrical lens enables scanning a light plane, but does not provide a mechanism for controlling scanning speed. Consequently, the scanning rate of the light plane is faster than the capture rate of the event camera. We overcome this problem by using a laser that we pulse at a controllable frequency. If this frequency is the same as that of the ultrasound transducer, and assuming laser pulsation starts at $t=0$, the light intensity at the focal plane will become:
\begin{align}
    I(x, t) = \sum_{k,l} \delta(x + k\lambda_\us - l \underbrace{c_\us T_\us}_{\lambda_\us}) \delta(t \bmod T_\us), 
    \label{eq:pulsed_intensity_focalplane}
\end{align}
where $l$ indexes laser pulses, and the inter-pulse time $T_\us = f_\us^{-1}$ equals the ultrasound period. \Cref{eq:pulsed_intensity_focalplane} describes a pulse train that does not translate in space but flickers in time. Every $l^{\text{th}}$ laser pulse illuminating the $l^{\text{th}}$ period of the ultrasonic wave will generate a line at the origin ($x = 0$).
 
To move the lines spatially, we instead pulse the laser at a frequency slightly offset from the ultrasonic one. If $\alpha$ is the ratio of ultrasonic to laser frequency, \cref{eq:pulsed_intensity_focalplane} becomes:
\begin{align}
    I(x, t) = \sum_{k,l} \delta(x + k\lambda_\us - l \alpha \lambda_\us) \delta(t \bmod \alpha T_\us). 
    \label{eq:steerable_focalplane}
\end{align}
Therefore, when the $l^{\text{th}}$ laser pulse illuminates the $l^{\text{th}}$ period of the ultrasonic wave, the focused line will appear at $x = (1-\alpha)l \lambda_\us$ making the line move spatially from pulse to pulse. The line frequency is equal to the beat frequency $(1-\alpha)f_\us$ between ultrasound and laser.

Besides serial line scanning, our laser pulsing design allows us to place lines at a non-serial set of locations $x_l;\, l \in \{1, 2, \dots, L\}$, by emitting laser pulses at times 
\begin{align}
    T_l = \frac{x_l + l\lambda_\us}{l c_\us}. 
    \label{eq:non_serial_locations}
\end{align}
This capability facilitates adaptive scanning of regions of interest, and in turn yet faster structured light rates (\cref{sec:results}).

The above analysis assumed a collimated beam. If we use a diverging beam, the line scanning behavior remains the same, but the distance between the lines will increase. In our prototype, we use a diverging beam to increase the field of view of the structured light system (\cref{fig:schematic}). 

\paragraph{Comparison to \citet{pediredla2023megahertz}.} Our acousto-optic laser line scanning system is inspired by that introduced by {\protect\NoHyper\citeauthor{pediredla2023megahertz}\protect\endNoHyper}, with two important differences:
\begin{enumerate*}
    \item {\protect\NoHyper\citeauthor{pediredla2023megahertz}\protect\endNoHyper} focus on point scanning for lidar applications, whereas we focus on line scanning for structured light. Therefore, their system uses two linear transducers, whereas ours using just one, effectively replacing spherical GRIN waveguides with cylindrical ones.
    \item {\protect\NoHyper\citeauthor{pediredla2023megahertz}\protect\endNoHyper} control the position of the focused spot by modulating the phase of the ultrasound, whereas we control the position of the focused line by pulsing the laser. The approach of {\protect\NoHyper\citeauthor{pediredla2023megahertz}\protect\endNoHyper} creates two challenges: First, it requires using an expensive broadband RF amplifier (ENI-300L, USD \num{23000}), as the ultrasonic wave is no longer a monotonic sinusoid. Second, it requires that the transducers operate at non-resonant frequencies, where they consume more power. By contrast, our approach allows using a much less expensive narrowband amplifier (USD \num{250}) and operating at lower power. 
\end{enumerate*}

\paragraph{Structured light.} For structured light, we use our AO light scanning device together with an event camera to form a stereo pair. We orient the planar transducer of our device such that the projected light planes are as orthogonal to the corresponding epipolar lines on the event camera as possible. By synchronizing the AO device and event camera, we can perform structured light scanning using the classical swept-plane procedure \citep{lanman2009build,bouguet19993d}, which we visualize in \cref{fig:structured_light}: As the AO device scans a light plane, light reflected off the scanned object forms a vertical curve on the image plane that sweeps horizontally across the field of view, triggering sparse events (ideally one event per sensor row for each scan position, assuming perfect optics). From each such event, we can reconstruct a depth value for the corresponding pixel via triangulation, by backprojecting a ray and intersecting it with the light plane that triggered the event.

    \begin{figure*}
    \centering
    \includegraphics[width=\linewidth]{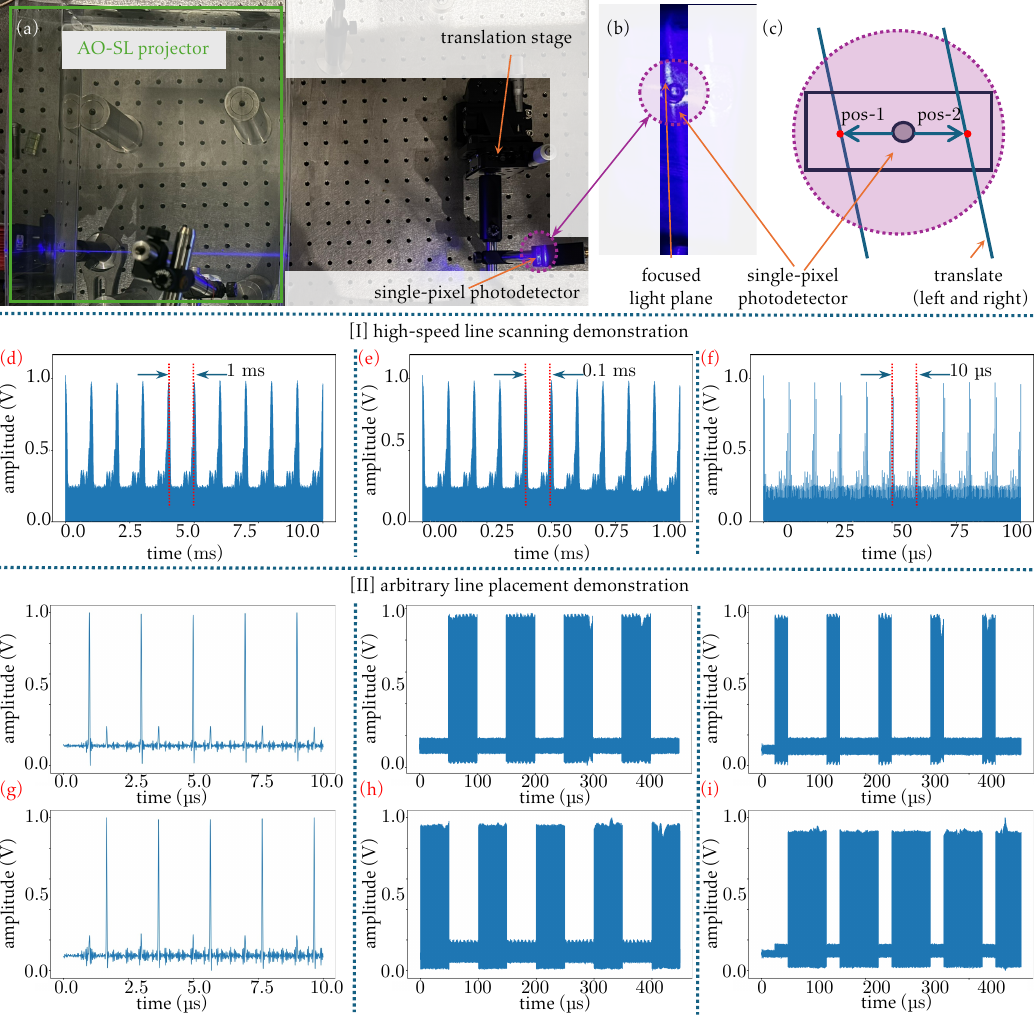}
     \caption{{\textbf{Experimental validation of [I] high-speed line scanning and [II] arbitrary line placement.}
    Our AO device is capable of translating the projected line at high speeds and positioning the lines at arbitrary spatial locations.
    (a) To validate these capabilities, we place a fast single-pixel photodetector (PD) mounted on a translation stage in the scene and monitor its output using an oscilloscope.
    (b) We show the line focused on the single-pixel photodetector surface.
    (c) We place the PD at two arbitrary positions (pos-$1$ and pos-$2$) where we intend to place arbitrary lines. This is equivalent to placing two PDs and monitoring their outputs.
    \textbf{[I]. High-speed scanning:} To demonstrate temporal scanning capability, the photodetector is fixed at a static location while the AO device scans the line periodically across it. (d)---(f) show oscilloscope outputs at scanning speeds of \SI{1}{\kilo\fps}, \SI{10}{\kilo\fps}, and \SI{100}{\kilo\fps}. The periodic peaks and their consistent temporal spacing confirm successful scanning at the target frequencies.
    \textbf{[II]. Arbitrary line placement:} To demonstrate arbitrary line placement, we place the PD at pos-$1$ or pos-$2$ and vary the line placement strategy. In (g)---(i), the top and bottom row show the oscilloscope outputs for pos-$1$ and pos-$2$, for various line placements: In (g), alternate lines are directed to pos-$1$ and pos-$2$, resulting in alternating signal peaks. In (h), $100$ lines are directed to pos-$1$ followed by $100$ lines to pos-$2$ producing a square waveform at \SI{10}{\kilo\hertz}. In (i), a sequence of $50$ lines at pos-$1$ and $150$ lines at pos-$2$, results in a waveform with a duty cycle of \SI{25}{\percent} at pos-$1$ and \SI{75}{\percent} at pos-$2$.
    }}
    \label{fig:photodetector_adaptive_scanning}
    \label{fig:photodetector_high_fps}
\end{figure*}

    \section{Hardware prototype}
\label{sec:prototype}

In this section, we detail hardware components and the system implementation. \Cref{fig:schematic} shows a schematic of the optical setup, and \cref{fig:hardware_prototype} shows our hardware prototype. 

\paragraph{Light scanning device.}
To implement the acousto-optic light scanner, we use an ultrasonic transducer of size $\qty{12.5}{\milli\meter} \times \qty{12.5}{\milli\meter}$ tuned to operate at a $\qty{2}{\mega\hertz}$ frequency. 
We drive the transducer using a Siglent SDG 1032X arbitrary-waveform generator (AWG). We amplify the AWG output using a narrowband RF amplifier (rated at \qty{50}{\watt} and operating frequency \qtyrange{1}{3}{\mega\hertz}) before applying it to the transducer. 
For illumination, we use a Thorlabs NPL45B pulsed laser that we synchronize and drive using the AWG to achieve line scanning. In the supplementary document, we provide detailed descriptions of the circuitry and the associated waveforms used for static and controlled steering of light planes, along with visualizations of their effects on illumination within the scanning system. We use a Thorlabs LA1274 \qty{40}{\milli\meter} convex lens to diverge the beam enough to cover the scanned scene. For comparison with previous steering techniques, we include in our system a Thorlabs GVS211 single-axis Galvo mirror, which we keep stationary when we use our acousto-optic light scanning.

\paragraph{Event camera.} We use the Prophesee EVK-4 event camera, equipped with Sony's IMX636 sensor. This sensor can process up to \SI{1}{\giga Events \per \second}, with events recorded at a temporal resolution of \qty{1}{\micro\second} and spatial resolution of  $1280 \times 720$ pixels. As the horizontal axis has the fastest readout and provides timestamping, we orient this axis parallel to the light plane. 
This alignment allows us to utilize the event camera's full bandwidth~\cite{muglikar2021esl}, corresponding to a maximum achievable frame rate of approximately $\nicefrac{10^6}{720} = \SI{1388}{\hertz} \approx \SI{1}{\kilo \hertz}$.
We replace the default lens of the EVK-4 camera with a Rokinon DS50M-C full-frame lens of focal length $\SI{50}{\milli\meter}$ and speed $f/1.5$ to improve light efficiency during scanning.

We observed empirically that, with default settings, the event camera fails to capture events when sweeping light planes at frequencies higher than \SI{100}{\hertz}. To mitigate this problem, {we follow \citet{gallego2020event}} and set the camera's parameters \texttt{bias-hpf} and \texttt{bias-refractory} to their maximum values and remove the \texttt{bias-off} parameter, allowing a single polarity to utilize the complete bandwidth of the camera. 

\paragraph{Synchronization.} To synchronize events with the light planes that trigger them, one option is to use the event camera's hardware trigger mechanism and synchronize with the planes' timestamps. However, at high-speed scanning when the event camera generates several events, we observed that this mechanism fails. We thus develop a software-based synchronization system: We project light planes continuously and record them with the event camera. We then use the event camera's timestamp for the left-most line and the corresponding AO light plane's timestamp for synchronization.  

\paragraph{Geometric calibration.} As our light scanning device is highly repeatable, we precalibrate scanned light planes through their linear intersections with two orthogonal reference planes, which we create using two LCD screens: We display blinking checkered patterns on the screens to calibrate the reference planes with respect to the event camera, then keep them turned off during structured light. We visualize this arrangement in the supplement. The procedure uses standard geometric calibration algorithms \citep{lanman2009build,bouguet19993d}.

\paragraph{Validation of high-speed line scanning.}
To validate that our AO system can achieve line scanning speeds exceeding \SI{1}{\kilo\hertz}, we use a fast single-pixel photodetector (Thorlabs DET25A, bandwidth \SI{2}{\giga\hertz}), as in \cref{fig:photodetector_high_fps}: We position the photodetector directly in front of the AO system, so that the detector will produce a sharp temporal response whenever the line sweeps the detector's active area. We use a high-speed oscilloscope to read out the detector response. Each instance of the line intersecting the detector results in an intensity peak in the detector response, and the temporal separation between these peaks equals the line scanning period. We show oscilloscope traces for three target scanning speeds, \SI{1}{\kilo\fps}, \SI{10}{\kilo\fps}, and \SI{100}{\kilo\fps}; the traces confirm that our AO system achieves these speeds(\cref{fig:photodetector_high_fps}\textcolor{red}{(d---f)}).

\paragraph{Validation of arbitrary line scanning.} To validate that our AO system can place all lines at target locations, we placed the fast single-pixel photodetector on a translation stage, as in \cref{fig:photodetector_adaptive_scanning}. We illuminated two lines in three different sequences:

\begin{enumerate*}
    \item alternately illuminate each line; 
    \item illuminate each line 100 times before alternating; 
    \item illuminate first line 50 and second line 150 times. 
\end{enumerate*}

The photodetector output confirms that our system achieves the target scanning sequences((\cref{fig:photodetector_high_fps}\textcolor{red}{(g---i)}).

    \begin{figure*}[t]
    \centering
    \includegraphics[width=\linewidth]{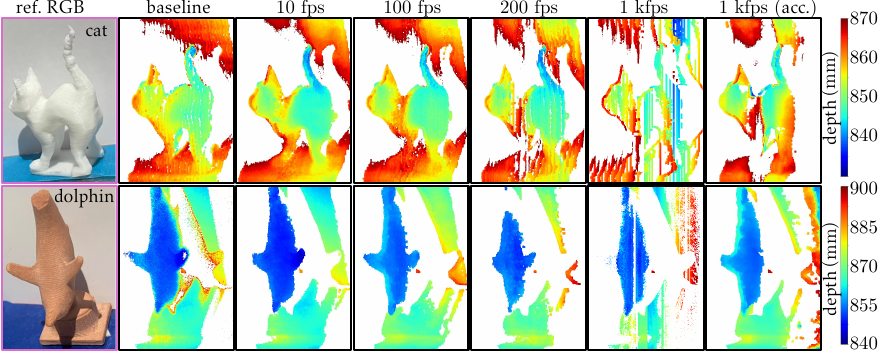}
    \caption{\textbf{Depth scanning of static scenes at different scan rates}. We scan a cat (top) and a dolphin (bottom) figurine, using the proposed AO SL system, at scan rates of \SI{10}{\fps}, \SI{100}{\fps}, \SI{200}{\fps}, and \SI{1}{\kilo\fps}. We use Galvo-based SL at \SI{10}{fps} as the baseline for comparison. 
    In \cref{tab:metrics}, we quantify reconstruction quality. Our proposed AO SL system reconstructs depth with high fidelity, but quality worsens as frame rate increases. At high frame rates, the event camera drops events randomly as the number of events exceeds its readout bandwidth. Future event cameras with higher bandwidth can mitigate this problem. We emulate them by scanning the static scene several times and accumulating the events to account for dropped events.}
    \label{fig:static_results}
\end{figure*}
    \begin{figure*}[t]
  \centering
  \includegraphics[width=\textwidth]{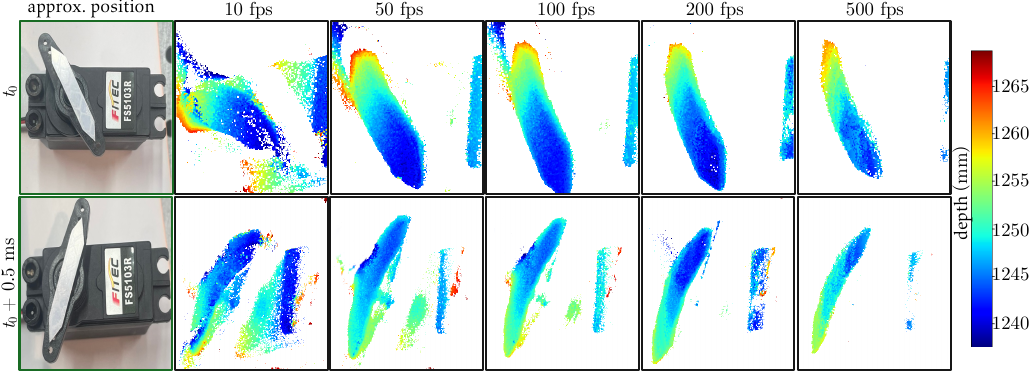}
  \caption{\textbf{Depth scanning of a servo motor rotating at \SI{70}{\rpm}}. Each row displays two frames (corresponding to times $t_0$ and $t_0 + \qty{0.5}{s}$) from scans obtained at scan rates of \SI{10}{\fps}, \SI{50}{\fps}, \SI{100}{\fps}, \SI{200}{\fps}, and \SI{500}{\fps}. At \SI{10}{\fps}, the scan does not produce a correct depth map, as the rapid motion in the scene exceeds the scan rate. At \SI{50}{\fps}, the scan provides a depth map that is more accurate but still exhibits motion blur (visible more clearly in the supplementary video). As scan rates increase to \SI{500}{\fps}, the depth maps become progressively more accurate. 
  }
  \label{fig:dynamic_servo_results}
\end{figure*}

    \begin{table}[t]
\centering
\caption{\textbf{Impact of scan rate.} (CD: chamfer distance in \SI{}{\milli\meter}; F1: F1-score; Pr: precision; Re: recall.)}\label{tab:metrics}
\begin{tabularx}{\linewidth}{@{} >{\hsize=0.8\hsize}X >{\hsize=1.5\hsize}X >{\hsize=1\hsize\centering\arraybackslash}X >{\hsize=0.9\hsize\centering\arraybackslash}X >{\hsize=0.9\hsize\centering\arraybackslash}X >{\hsize=0.9\hsize\centering\arraybackslash}X @{}}
\toprule
    & fps & CD ($\downarrow$) & Pr ($\uparrow$) & Re ($\uparrow$) & F1 ($\uparrow$)\\
    \midrule
       \multirow{5}{*}{cat} & 10 & 1.48 & 0.762 & 0.910 & 0.830 \\
                         & 100 & 1.25 & 0.812 & 0.953 & 0.877 \\
                         & 200 & 1.67 & 0.738 & 0.956 & 0.833 \\
                         & 1000 & 4.56 & 0.146 & 0.981 & 0.253 \\
                         & 1000 (acc.) & 4.30 & 0.524 & 0.827 & 0.641 \\
    \midrule
    \multirow{5}{*}{dolphin} & 10 & 0.750 & 0.958 & 0.939 & 0.949 \\
                             & 100 & 1.273 & 0.961 & 0.817 & 0.883 \\
                             & 200 & 1.583 & 0.915 & 0.803 & 0.855 \\
                             & 1000 & 2.342 & 0.844 & 0.687 & 0.758 \\
                             & 1000 (acc.) & 1.338 & 0.986 & 0.926 & 0.955 \\
    \bottomrule
\end{tabularx}
\end{table}

    \begin{figure*}[t]
    \centering
    \includegraphics[width=\linewidth]{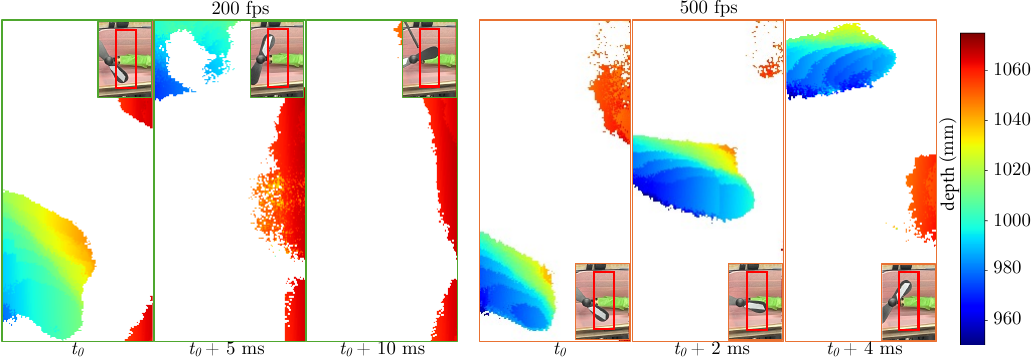}
    \caption{\textbf{Depth scanning of a fan rotating at \qty{1800}{\rpm}}. The left column displays three depth frames captured at a scan rate of \SI{200}{\fps}. The right column shows the same for a higher scan rate of \SI{500}{\fps}. The depth maps show significantly more motion blur on the blade area at \SI{200}{\fps} than at \SI{500}{\fps}. The red regions are from the background calibration plane. The insets show the approximate orientation of the fan blade, and the area inside the red box is the scanned region.}
    \label{fig:dynamic_fan_results}
\end{figure*}

    \section{Experiments and analysis}\label{sec:results}

In this section, we systematically evaluate the performance of our proposed structured light system on static and dynamic scenes. We show that our system can reconstruct 3D scenes at the maximum frame rate of the event camera at megapixel resolution. We also show an adaptive 3D scanning method that allows to selectively scan parts of the scene, thereby achieving a $10\times$ higher frame rate. We empirically show that structured light systems using Galvo mirrors cannot effectively perform adaptive 3D scanning. 

\paragraph{Supplement.} The supplement shows additional experimental and other results, including:
\begin{enumerate*}
    \item A video explaining our method.
    \item Videos showing temporal sequences of depth scans at various scan rates.
    \item A PDF document and short videos detailing setup implementation and calibration.
\end{enumerate*}

\subsection{Static scene results}

We first scan static scenes, to quantitatively and qualitatively evaluate the effect of scan rate on reconstruction performance. In \cref{fig:static_results}, we use two figurines as test objects for scanning. We first used Galvo-based SL at a slow speed (\SI{10}{\fps}) to compute ground truth for comparison. We then used the AO device to scan the object at \SI{10}{\fps}, \SI{100}{\fps}, \SI{200}{\fps}, and \SI{1}{\kilo\fps} scan rates. The figure shows the captured depth maps, and \cref{tab:metrics} provides quantitative metrics for reconstruction quality. In the supplement, we provide 3D visualizations of the reconstructed point clouds. 

From \Cref{fig:static_results}, we observe that our system produces depth results nearly identical to those from the Galvo-based system at low scanning rates. As scanning rate increases, reconstruction quality progressively deteriorates due to two reasons:
\begin{enumerate*}
    \item Higher scan rates result in decreased exposure time and signal-to-noise ratio, in turn causing a lot more noise events to be triggered.
    \item The linewidth from the AO device is around 10 pixels, thus each line triggers events at multiple sensor columns. At high scan rates, this increased number of events saturates the camera, making it randomly lose some events (columns) and causing the missing stripes. 
\end{enumerate*}
At \SI{1}{\kilo\fps}, the number of missing columns due to the second issue becomes significant. To mitigate this issue, we experimented with accumulating multiple frames (``acc.'' in \cref{fig:static_results,tab:metrics}), effectively simulating a higher-bandwidth camera. We observe that doing so produces complete depth scans. This effect is scene-dependent, and sparse scenes as in \cref{fig:teaser} do not suffer from reduced reconstruction quality due to limited readout bandwidth. 

In \Cref{tab:metrics}, we quantify the 3D reconstruction fidelity using chamfer distance, precision, recall, and F1 score. As the frame rate increases, all metrics deteriorate as expected. With accumulation, the metrics improve significantly. 
  
\subsection{Dynamic scene results}

\begin{figure}
    \centering
    \includegraphics[width=\linewidth]{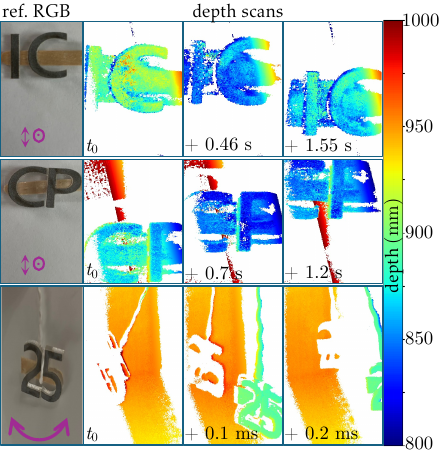}
     \caption{\textbf{Non-periodic dynamic scenes}: Depth scans of characters \enquote{IC}, \enquote{CP}, and \enquote{25}, covered with retroreflecting tape. In the top and middle rows, the characters are mounted on a stick and moved rapidly in a top-to-bottom (\textcolor{purple}{$\updownarrow$}) sweeping motion, with some additional forward-backward translation (\textcolor{purple}{$\odot$}). In the bottom row, the characters are swung by hand using a string. For each row, the RGB image to the left shows the scene and visualizes the approximation motion trajectory. The images to the right show the depth frames at different points of the trajectory, captured at \SI{180}{\fps} (top and middle row) or \SI{200}{\fps} (bottom row). The supplement includes a video of continuous depth scanning as the targets move.}
    \label{fig:iccp_fig}
\end{figure}

To showcase the speed and characterize the performance of our structured light system, we use two types of dynamic scenes:
\begin{enumerate*}
  \item Periodic dynamic scenes, including a servo motor rotating at \qty{70}{\rpm} (\cref{fig:dynamic_servo_results}) and a high-speed fan rotating at \qty{1800}{\rpm} (\cref{fig:dynamic_fan_results}).
  \item Non-periodic dynamic scenes, including zig-zag motion and left-right oscillation (\cref{fig:iccp_fig}).
\end{enumerate*}

\paragraph{Periodic dynamic scenes.} We use these scenes to perform experiments with controllable motion and speeds, to assess motion blur and determine optimal scan speeds.
    
For the \qty{70}{\rpm} servo (blade diameter \SI{35}{\milli\meter}), \cref{fig:dynamic_servo_results} shows results for five scan rates: \SI{10}{\fps}, \SI{50}{\fps}, \SI{100}{\fps}, \SI{200}{\fps}, and \SI{500}{\fps}. The servo is rotating clockwise, and we show depth scans of two frames separated by \qty{0.5}{s}. As the fan blades are dark and we are scanning at very low exposures, we used retroreflective tape to increase reflectivity. 

The RGB images show the orientation of the blade and the region where the reflectors are located. At \SI{10}{\fps}, the depth maps suffer from significant blur. At \SI{50}{\fps}, the depth maps improve significantly but still have motion blur (seen as increased blade thickness). The results become increasingly sharp as the scan rate increases. The video results in the supplement show continuous depth scans of the dynamic scene at all scan rates. 

For the more challenging fan rotating at \qty{1800}{\rpm} (blade diameter \SI{72}{\milli\meter}), \cref{fig:dynamic_fan_results} shows results at scan rates of \SI{200}{\fps} and \SI{500}{\fps}. 

The fan blade has considerably more motion blur at \SI{200}{\fps} than at \SI{500}{\fps}. The former is representative of the expected performance of the prior state of the art (\citet{dashpute2023event}, who reported scan rates of \SI{250}{\fps}).

\paragraph{Non-periodic dynamic scenes.} In \cref{fig:iccp_fig}, we show results from experiments using hand-held objects undergoing various types of motion. Our system successfully acquires full-frame depth despite the fast motion. The supplement shows the full depth videos we captured with our system.

\subsection{Adaptive depth scanning}
\label{sub_sec:Adaptive_depth_scanning}

As we mentioned in \cref{sec:prototype}, the event sensor can capture only up to \qty{1}{k} frames per second at megapixel resolution. In a dynamic scene, typically depth changes only at few pixels corresponding to the moving objects. Therefore, one way to increase scan rate beyond the sensor's theoretical full-frame rate is to perform adaptive scanning, where we only scan the regions where the depth has changed. \Citet{muglikar2021event} have previously demonstrated that adaptive scanning of only regions of interest improves spatial and temporal resolution. As in their work, we can illuminate only the important regions of the scene to reduce the event camera bandwidth and increase overall scan rate. 

Our AO light scanning device can illuminate any arbitrary line at the same high rate (ultrasound frequency) as serial scanning, making it well-suited for adaptive scanning. Importantly, using \cref{eq:non_serial_locations}, our AO device can perfectly distribute the full power of the light source at \emph{only} the lines of interest (i.e., it is a \emph{redistributive} line projector \citep{o2015homogeneous}). By contrast, using a Galvo mirror to perform adaptive scanning is slow, as the mirror is no longer operating at resonant mode. Additionally, due to the need for continuous mechanical rotation from one target line to another, a Galvo mirror ends up additionally illuminating the region between target lines (i.e., it is \emph{not} a redistributive projector). This deficiency not only wastes light power, but also generates extraneous events, making  adaptive scanning overall less effective.

\begin{figure}[t]
  \centering
  \includegraphics[width=\linewidth]{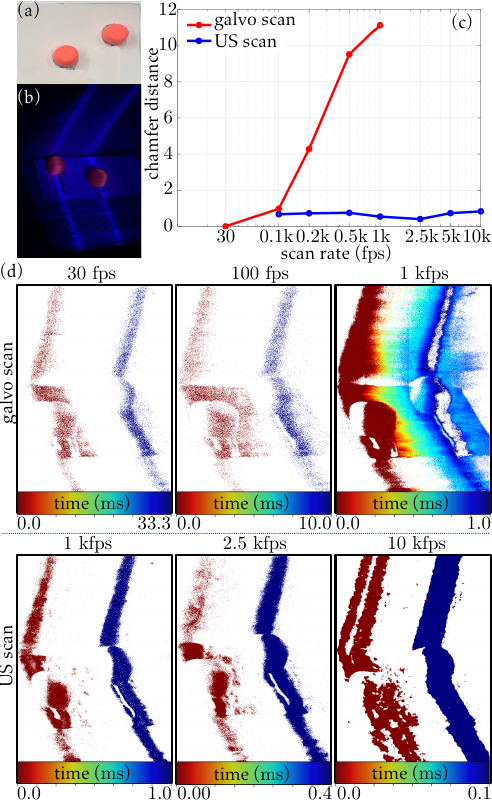}
  \caption{\textbf{Adaptive scanning} of two regions of interest (orange knobs) in a scene shown in (a). We aim to illuminate and 3D scan only the knobs (b), using both the Galvo mirror and our AO device.  In (c), we show the depth error measured for these knobs when we scan with the two devices. At low scan rates, the Galvo mirror is effective at adaptively scanning only the knobs, but its performance deteriorates as the scan rate increases: Due to the need to mechanically rotate the beam from one location to another, the Galvo mirror must also illuminate the region between the knobs. At high scanning speeds, the fraction of total scanning time spent illuminating this region increases, decreasing performance. By contrast, our AO device can illuminate only the knobs, resulting in effective adaptive scanning even at \SI{10}{\kilo\fps}.}
  \label{fig:adaptive_scanning}
\end{figure}

To demonstrate adaptive scanning, we show in \cref{fig:adaptive_scanning} a scene comprising two orange knobs at two different depths. We scanned the positions of only these knobs with the Galvo mirror device and the AO device at various scan rates. 
From the plot in \cref{fig:adaptive_scanning}, we observe that the Galvo mirror device produces accurate depth at low frame rates. However, as the frame rate increases, the depth error increases significantly. By contrast, the accuracy of the AO device remains fairly constant, even at a very high scan rate. 

To better understand this performance difference, we show captured event data in \cref{fig:adaptive_scanning}(b), with color representing the time stamp of the events. 
In the supplement, we provide the event stream videos to show the behavior more clearly. We also detail the light pulse timing control used to achieve controlled oscillation between the desired light planes at specified \SI{}{\fps} in the supplementary material.
From this data, we observe that at low scan rates, the Galvo mirror device projects most of the laser's light output to the two lines of interest, but as the scan rate increases, it projects most of the light output incorrectly to the region between the two lines. By contrast, the AO device always projects all light at only the two lines of interest. 

We note that we could not scan the Galvo mirror device beyond a \SI{1}{\kilo\hertz} rate, as that is the physical limit for these devices. For our AO device, we could theoretically go up to \SI{1}{\mega\hertz} to scan two lines (given that the transducer operates at \SI{2}{\mega\hertz}). However, as the frame rate increases, the amount of light decreases, and we noticed empirically that at \SI{20}{\kilo\hertz} the amount of light reaching the event sensor is below its detection threshold. Therefore, though  theoretically we can scan up to \SI{1}{\mega\hertz} adaptively, practically, we are limited by our laser power to a \SI{10}{\kilo\hertz} scan rate, which is still $10 \times$ higher than the event camera's full-frame frame rate.

    \section{Discussion and conclusion}
\label{sec:discussion}

We designed an acousto-optic device to scan light planes at unprecedented rates of up to $\SI{2}{\mega\hertz}$, and combined it with an event camera to perform swept-plane structured light. This combination allowed us to achieve full-frame 3D scanning at $\SI{1000}{\fps}$, 3--4$\times$ faster than the state of the art \cite{dashpute2023event}. However, our method has a few limitations, some of which can be addressed by modifications to the prototype, whereas others point towards future research directions.

On the first front, similar to \citet{pediredla2023megahertz}, our prototype has a small aperture that, due to diffraction, results in large linewidths and thus reduced spatial and depth resolution. Using a transmission medium where the speed of sound is faster than in water would increase the ultrasound wavelength and hence the aperture size, addressing this issue. Additionally, our prototype uses a laser with low emission power ($\approx \SI{1}{\milli\watt}$), necessitating light accumulation over multiple scans at higher frame rates, as well as the use of a large aperture lens that reduces depth of field. Using a better laser would mitigate both issues. Lastly, our current prototype has a limited field of view of $\SI{38}{\milli\meter}$, constraining the size of objects that can be scanned. This issue can be mitigated using better optics (for example, an F-theta lens), at the cost of increased linewidth. Lastly, our prototype works by modulating a large water volume,

{In terms of more fundamental limitations, our prototype requires modulating a large volume of an optically transparent medium (e.g., water), resulting in a large form factor that is impractical for miniaturization. Enabling miniaturization requires research on realizing the light-focusing behavior of our prototype using standard acousto-optic modulators.} Additionally, the limited bandwidth of the event camera is the bottleneck that prevents faster 3D scanning rates, creating a gap of three orders of magnitude with the light scanning rate. Closing this gap requires research into sensors that not only have drastically larger readout bandwidths, but also can operate at drastically shorter exposure times. Single-photon avalanche diodes are a promising option to this end, as they are suitable for ultra-wideband operation \citep{wei2023passive} and structured light under minuscule exposures \citep{sundar2022single}. Additionally, our system creates opportunities for combinations with other imaging modalities that use additional dimensions of light (e.g., polarization, wavelength) to bring robustness against effects such as specularity, glare, and scattering that are challenging for structured light systems---including ours. Lastly, incorporating learning-based depth priors and reconstruction methods (e.g., diffusion, neural rendering) into our pipeline could further enhance reconstruction fidelity and robustness, especially under noisy or sparse event conditions.

  \ifpeerreview \else
  \section*{Acknowledgments}
This work was supported by the National Science Foundation under awards 2047341, 2107454, 2326904, and 2403122, as well as a Sloan Research Fellowship for Ioannis Gkioulekas. We thank Aniket Dashpute (Rice University) and Manasi Muglikar (University of Zurich) for their valuable insights on event camera parameters. We also thank Ziyuan (Quinton) Qu and Sarah K. Friday (Dartmouth College), for their assistance in setting up the experimental scenes.

  \fi

  \bibliographystyle{IEEEtranN}
  \bibliography{main}

  \ifpeerreview \else
  \begin{IEEEbiography}
    [{\includegraphics[width=1in,height=1.25in,clip,keepaspectratio]{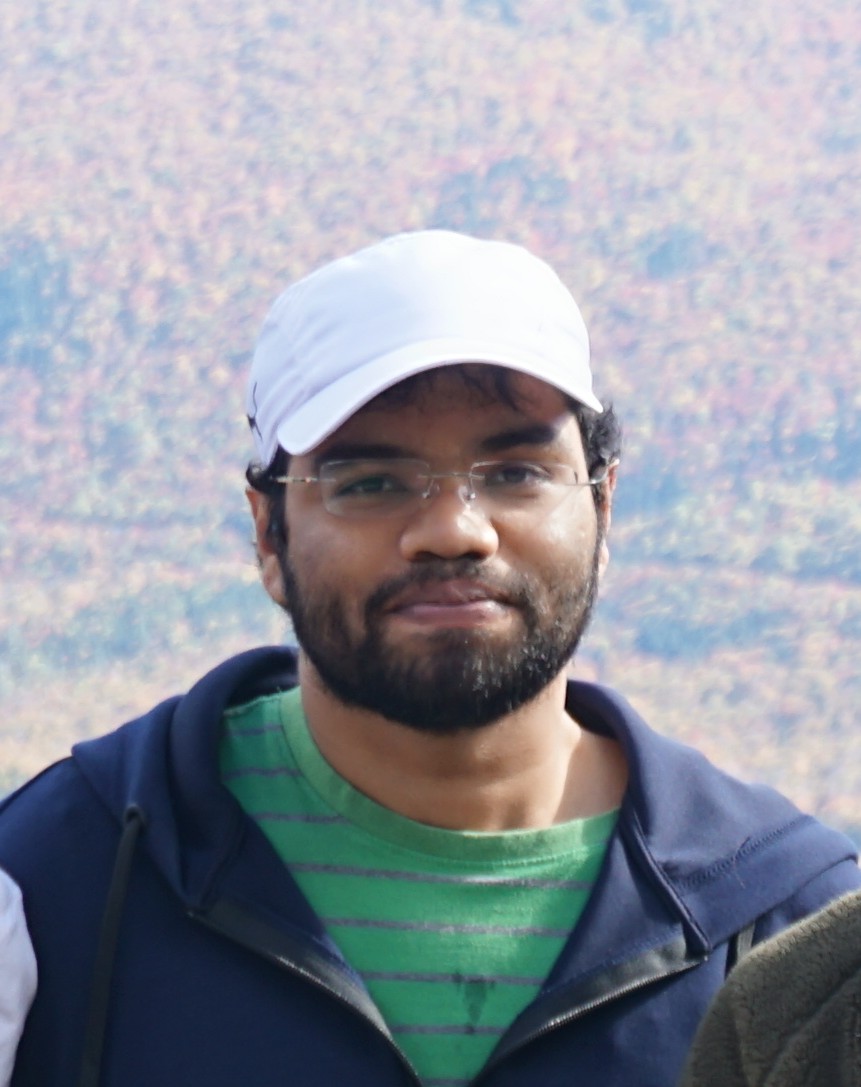}}]{Dhawal Sirikonda}is a Ph.D. student at Dartmouth College and a member of the Rendering and Imaging Science (RISc) Lab, advised by Prof. Adithya Pediredla. He works on problems in computational imaging. He earned his Master’s degree from IIIT Hyderabad, where he worked with Prof. P. J. Narayanan at the intersection of 3D vision and real-time graphics.
\end{IEEEbiography}

\begin{IEEEbiography}
    [{\includegraphics[width=1in,height=1.25in,clip,keepaspectratio]{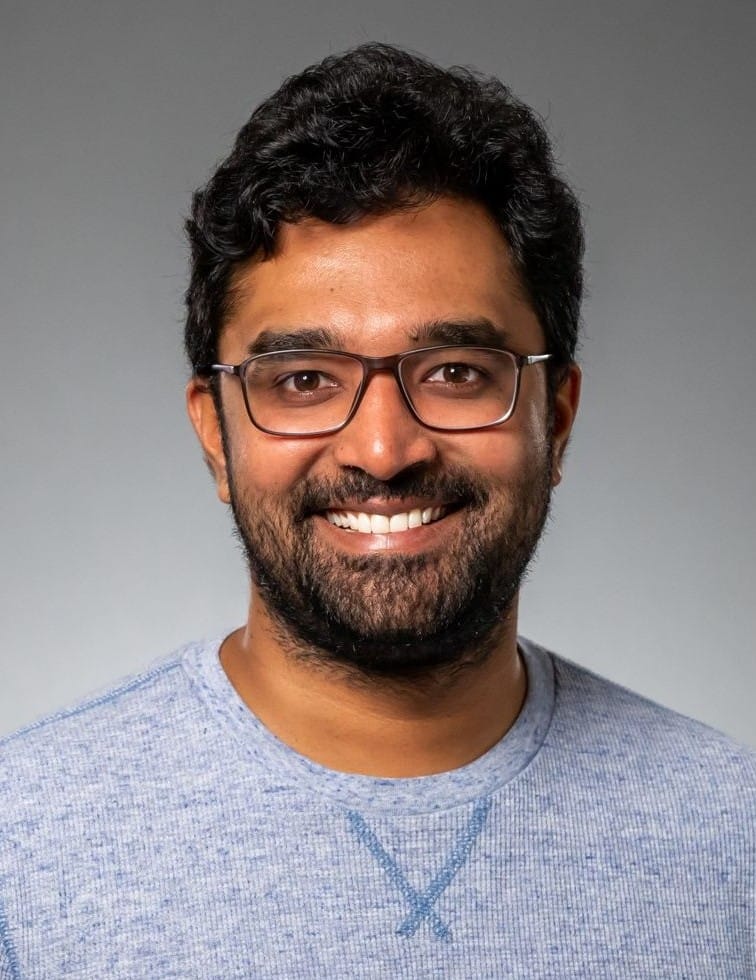}}]{Praneeth Chakravarthula}is an Assistant Professor at the University of North Carolina at Chapel Hill. His research focuses on novel computational imaging and display systems, at the intersection of optics, perception, computer graphics, optimization, and machine learning. Dr. Chakravarthula is a Senior Member of IEEE and Optica, recipient of the IEEE VR Best Dissertation Award, as well as multiple Best Paper and Demo Awards at premier venues such as ACM SIGGRAPH, IEEE VR, and ISMAR. He completed his postdoctoral research at Princeton University, received his Ph.D. from UNC Chapel Hill, and holds B.Tech and M.Tech degrees in Electrical Engineering with a specialization in Signal Processing from the Indian Institute of Technology (IIT) Madras. 
\end{IEEEbiography}

\begin{IEEEbiography}
    [{\includegraphics[width=1in,height=1.25in,clip,keepaspectratio]{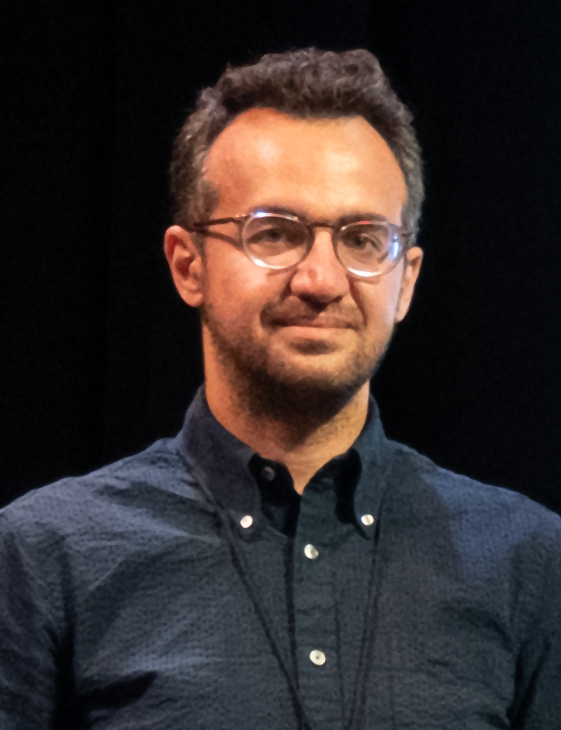}}]{Ioannis Gkioulekas}is an Associate Professor in the Robotics Institute at Carnegie Mellon University. He received his Ph.D. and M.S. in Engineering Sciences from Harvard University, and a Diploma in Electrical and Computer Engineering from the National Technical University of Athens. His research focuses on computational imaging and physics-based rendering. He is a recipient of the NSF CAREER Award, a Sloan Research Fellowship, and Best Paper Awards at CVPR and SIGGRAPH. He has served on numerous program committees and editorial boards in computer vision and graphics, and co-organizes the ICCP Summer School on Computational Imaging.
\end{IEEEbiography}

\begin{IEEEbiography}
    [{\includegraphics[width=1in,height=1.25in,clip,keepaspectratio]{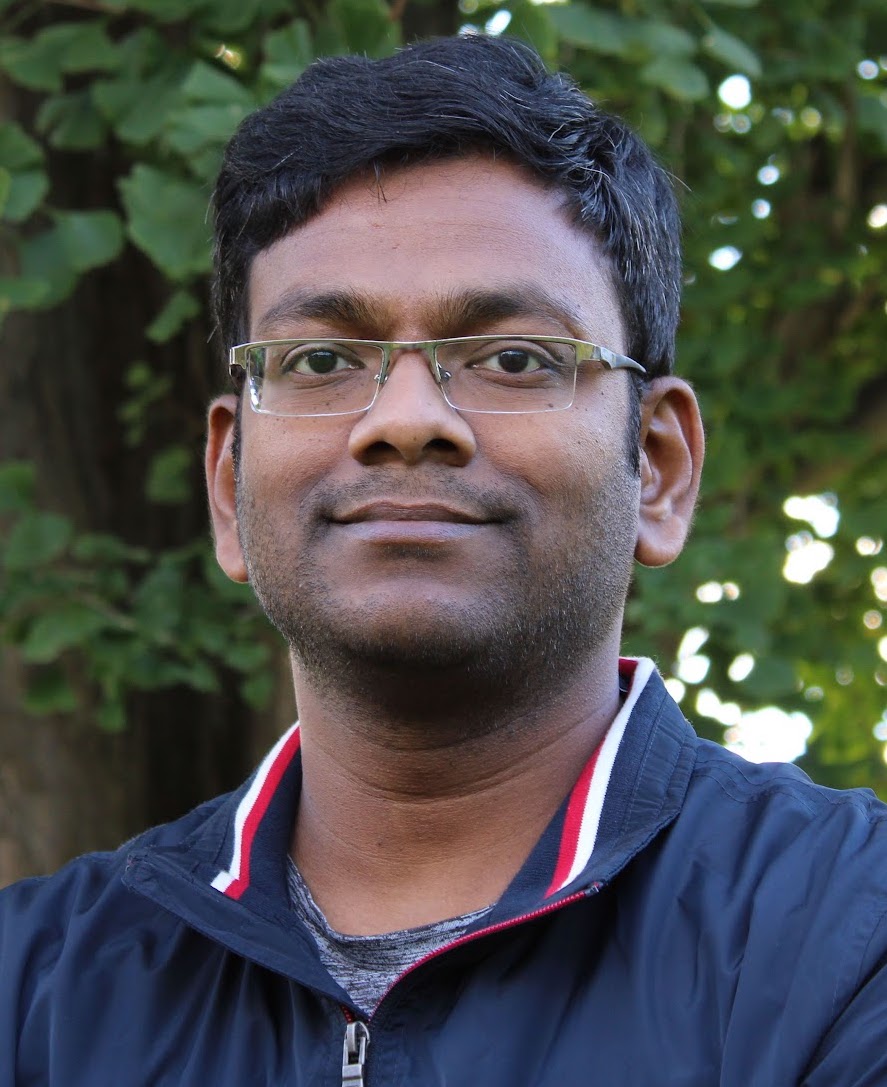}}]{Adithya Pediredla}is an Assistant Professor in the Department of Computer Science at Dartmouth College, where he leads the Rendering and Imaging Science (RISC) lab, focusing on the intersection of computer graphics, computational imaging, and computer vision. Before this role, he is a postdoctoral fellow at the Robotics Institute, Carnegie Mellon University. He has a Ph.D. from Rice University and a master’s degree from the Indian Institute of Science. He won the Ralph Budd Best Engineering Thesis Award for his Ph.D. thesis, the Prof. K. R. Kambati Memorial gold medal, and an innovative student project award from the Indian National Academy of Engineering for his Master’s thesis.
\end{IEEEbiography}

  \fi
\end{document}